\documentclass[conference]{IEEEtran}
\IEEEoverridecommandlockouts
\usepackage{cite}
\usepackage{amsmath,amssymb,amsfonts}
\usepackage{algorithmic}
\usepackage{graphicx}
\usepackage{textcomp}
\usepackage{xcolor}

\usepackage{booktabs} 
\def\BibTeX{{\rm B\kern-.05em{\sc i\kern-.025em b}\kern-.08em
    T\kern-.1667em\lower.7ex\hbox{E}\kern-.125emX}}
\begin{document}

\title{Does Bigger Mean Better? Comparative Analysis of CNNs and Biomedical Vision-Language Models in Medical Diagnosis}

\author{
\IEEEauthorblockN{Ran Tong}
\IEEEauthorblockA{\textit{University of Texas at Dallas}\\
United States\\
rxt200012@utdallas.edu}
\and
\IEEEauthorblockN{Jiaqi Liu }
\IEEEauthorblockA{\textit{Independent Researcher}\\
United States\\
jackyliu9747@gmail.com}
\and
\IEEEauthorblockN{Tong Wang}
\IEEEauthorblockA{\textit{Duke University}\\
United States\\
wangtongnellt@gmail.com}
\and
\IEEEauthorblockN{Xin Hu}
\IEEEauthorblockA{\textit{University of Michigan Ann Arbor}\\
United States\\
hsinhu@umich.edu}
\and
\IEEEauthorblockN{Su Liu}
\IEEEauthorblockA{\textit{Georgia Institute of Technology}\\
United States\\
sliu792@gatech.edu}
\and
\IEEEauthorblockN{Lanruo Wang}
\IEEEauthorblockA{\textit{University of Texas at Dallas}\\
United States\\
lxw220021@utdallas.edu}
\and
\IEEEauthorblockN{Jiexi Xu }
\IEEEauthorblockA{\textit{University of California, Irvine}\\
United States\\
xuj35513@gmail.com}

}

\maketitle

\begin{abstract}
The accurate interpretation of chest radiographs using automated methods is a critical task in medical imaging. This paper presents a comparative analysis between a supervised lightweight Convolutional Neural Network (CNN) and a state-of-the-art, zero-shot medical Vision-Language Model (VLM), BiomedCLIP, across two distinct diagnostic tasks: pneumonia detection on the PneumoniaMNIST benchmark and tuberculosis detection on the Shenzhen TB dataset. Our experiments show that supervised CNNs serve as highly competitive baselines in both cases. While the default zero-shot performance of the VLM is lower, we demonstrate that its potential can be unlocked via a simple yet crucial remedy: decision threshold calibration. By optimizing the classification threshold on a validation set, the performance of BiomedCLIP is significantly boosted across both datasets. For pneumonia detection, calibration enables the zero-shot VLM to achieve a superior F1-score of 0.8841, surpassing the supervised CNN's 0.8803. For tuberculosis detection, calibration dramatically improves the F1-score from 0.4812 to 0.7684, bringing it close to the supervised baseline's 0.7834. This work highlights a key insight: proper calibration is essential for leveraging the full diagnostic power of zero-shot VLMs, enabling them to match or even outperform efficient, task-specific supervised models.
\end{abstract}

\begin{IEEEkeywords}
Medical Imaging, Deep Learning, Vision-Language Models, Multimodal model, Zero-Shot Learning, Pneumonia Detection, Tuberculosis Detection, CNN
\end{IEEEkeywords}

\section{Introduction}
The automated analysis of medical images using deep learning has become a cornerstone of modern computational medicine, with applications ranging from oncology to radiology. One of the most critical diagnostic tasks is the interpretation of chest radiographs for conditions like pneumonia and tuberculosis, which require timely and accurate identification to ensure proper patient care. The importance of this problem has motivated the development of specialized models and curated datasets to advance the field.

This paper investigates the trade-offs between two distinct approaches for chest X-ray classification. The first is a supervised learning method using a lightweight Convolutional Neural Network (CNN). The second leverages the zero-shot capabilities of large-scale, pre-trained vision-language models (VLMs) that have been specialized for the medical domain. We apply these methods to two separate binary classification tasks: detecting pneumonia in the PneumoniaMNIST dataset \cite{b1} and identifying tuberculosis (TB) in the Shenzhen Chest X-ray dataset\cite{shenzhen_qims}.

Our methodology involves a direct comparison on both tasks. We first train a supervised CNN on each dataset to establish a strong baseline. We then evaluate a state-of-the-art medical VLM, BiomedCLIP \cite{b2}, in a zero-shot setting using descriptive text prompts. Our experiments reveal that the trained CNNs are formidable baselines in both scenarios. While the zero-shot performance of the VLM using a standard `argmax` approach is lower, we find that a simple calibration of the decision threshold on a validation set significantly boosts its performance. Notably, on the pneumonia task, a calibrated BiomedCLIP surpasses the supervised CNN. On the TB task, the same remedy dramatically improves the VLM's performance, making it highly competitive with the supervised model. This insight underscores a key challenge and a potential solution for deploying VLMs in clinical contexts.

The main contributions of this work are threefold:
\begin{enumerate}
    \item We demonstrate that a lightweight, task-specific CNN trained on standardized datasets is a highly competitive and resource-efficient baseline, even when compared against a large, pre-trained foundation model.
    \item We show that a state-of-the-art medical VLM, when applied in a default zero-shot setting using standard  classification, fails to outperform the supervised CNN baseline on these diagnostic tasks.
    \item We then demonstrate that a calibration protocol, which optimizes the decision threshold on a validation set, markedly improves the VLM's performance. This remedy elevates the VLM to be highly competitive with the supervised model for tuberculosis detection and superior for pneumonia detection, highlighting a crucial step for deploying zero-shot models effectively.
\end{enumerate}

\section{Related Work}
The application of deep learning to medicine has a rich and expanding history, with machine learning now integral to infectious disease prediction, diagnosis, and forecasting, as detailed in recent systematic reviews \cite{wang2025systematic}. A prominent area is medical imaging, where CNNs were adapted for tasks such as disease classification, localization, and segmentation \cite{b7,zhang2015thalamocortical}. Beyond imaging, machine learning has also been pivotal in predictive modeling using different data modalities. For instance, researchers have focused on forecasting Parkinson's disease progression from longitudinal biomarkers \cite{b6}, while others have developed methods to visualize non-linear relationships in genetic and microbiome data to better understand disease associations \cite{zhu2018nonlinear}. To facilitate standardized and reproducible research in imaging, benchmark datasets like MedMNIST were introduced, offering a collection of pre-processed medical image datasets for classification tasks \cite{b1}. Our work utilizes the PneumoniaMNIST subset of this collection.

More recently, the field has shifted towards leveraging large-scale, pre-trained models, particularly vision-language models (VLMs) that learn joint representations of images and text. The pioneering CLIP model demonstrated remarkable zero-shot learning capabilities by training on vast internet-scale data \cite{b8}, and its foundational architecture has proven highly adaptable to a wide range of specialized content analysis tasks, such as detecting harmful content \cite{b5}. This paradigm has been extended to specialized domains, a trend detailed in a recent review \cite{b4} (see also \cite{su2025applications}), which surveys the evolution of large language models and multimodal systems in medicine. This evolution is not only leading to domain-specific VLMs like BiomedCLIP \cite{b2}, but is also inspiring research into core challenges for medical AI. These challenges include understanding complex, structured documents analogous to electronic health records \cite{liu2025markuplanguagemodelingweb}, developing efficient lightweight models suitable for clinical deployment \cite{li2025ammkd}, exploring the potential of smaller, specialized language models through advanced prompt strategies \cite{wang2025applications}, and ensuring that foundation models perform fairly and ethically on tasks like medical image segmentation \cite{li2025achieving}.

Our work is positioned within this growing body of research that seeks to understand and apply these powerful models to practical problems. Our work differs by providing a focused, head-to-head comparison between a traditional supervised baseline and a state-of-the-art medical VLM on two distinct, well-defined tasks. Our primary contribution is the investigation of classification threshold calibration, demonstrating that this simple step is crucial for unlocking the full potential of these models in a zero-shot setting.

\section{Methodology}
Our methodology is designed to provide a clear comparison between a standard supervised learning approach and a zero-shot approach using a pre-trained vision-language model (VLM). We detail the datasets, the model architectures, the training and evaluation procedures, and our proposed calibration remedy.

\subsection{Datasets and Preprocessing}
We use two publicly available chest X-ray datasets for our experiments.

\subsubsection{PneumoniaMNIST} This dataset, from the MedMNIST v2 collection \cite{b1}, consists of 5,856 chest X-ray images categorized as "normal" or "pneumonia". The data is pre-split into training (4,708 images), validation (524 images), and test (624 images) sets. For our supervised CNN, images are treated as single-channel (grayscale), resized to 64x64 pixels, and normalized.

\subsubsection{Shenzhen Tuberculosis Dataset} This dataset contains 662 chest X-rays, each labeled as "normal" (326 images) or "tuberculosis" (336 images). We perform a stratified split of the data into training (60 \%), validation (10\%), and test (30\%) sets. Preprocessing for the CNN is identical to that for PneumoniaMNIST.

For the zero-shot VLM evaluation on both datasets, the original images are processed by the VLM's specific preprocessing pipeline, which converts them to RGB and resizes them to 224x224 pixels.

\subsection{Supervised CNN Baseline}
We implement a lightweight Convolutional Neural Network (CNN) as a supervised baseline. Input radiographs are converted to single-channel grayscale, resized to $64{\times}64$, and normalized. During training, we use light data augmentation (random horizontal flips and small affine transformations).

The network architecture, $f_\theta:\mathbb{R}^{1\times64\times64}\!\to\!\mathbb{R}^2$, consists of three sequential convolutional blocks followed by a two-layer MLP head. Each block applies a $3{\times}3$ convolution, a ReLU activation, and a $2{\times}2$ max-pooling operation. The complete architecture is summarized in Table \ref{tab:cnn_arch}.

\begin{table}[htbp]
\caption{Supervised CNN Architecture Summary}
\begin{center}
\begin{tabular}{@{}lccc@{}}
\toprule
\textbf{Layer} & \textbf{Configuration} & \textbf{Output Shape} & \textbf{Params} \\
\midrule
Input & -- & $1\times64\times64$ & -- \\
Conv1\,$\to$\,ReLU\,$\to$\,Pool & $1\!\to\!16$ filters, $3{\times}3$ & $16\times32\times32$ & $160$ \\
Conv2\,$\to$\,ReLU\,$\to$\,Pool & $16\!\to\!32$ filters, $3{\times}3$ & $32\times16\times16$ & $4{,}640$ \\
Conv3\,$\to$\,ReLU\,$\to$\,Pool & $32\!\to\!64$ filters, $3{\times}3$ & $64\times8\times8$ & $18{,}496$ \\
Flatten & -- & $4{,}096$ & -- \\
FC1\,$\to$\,ReLU & $4096\!\to\!64$ & $64$ & $262{,}208$ \\
FC2 (logits) & $64\!\to\!2$ & $2$ & $130$ \\
\midrule
\multicolumn{3}{r}{\textbf{Total Parameters}} & \textbf{285,634} \\
\bottomrule
\end{tabular}
\label{tab:cnn_arch}
\end{center}
\end{table}

The model is trained to minimize the cross-entropy loss,
\[
\mathcal{L}_{\text{CE}}
= -\,\frac{1}{B}\sum_{i=1}^{B} \log p\!\big(y_i\mid \tilde{\mathbf{x}}_i\big),
\]
where probabilities $p(y_i\mid \tilde{\mathbf{x}}_i)$ are derived from the softmax of the output logits. We use the AdamW optimizer with a learning rate of $10^{-3}$ and weight decay of $10^{-4}$. The final model checkpoint is selected based on the highest ROC AUC score achieved on the validation set.

\subsection{Zero-Shot VLM Evaluation}
We evaluate the prominent medical VLM BiomedCLIP \cite{b2}. The evaluation is performed in a zero-shot manner, meaning the model is not trained or fine-tuned on our target datasets.

\subsubsection{Text Prototype Generation}
For each class in each dataset, we create a set of five descriptive text prompts. For example, the "pneumonia" class is described with prompts like "a chest X-ray showing pneumonia," while the "tuberculosis" class uses prompts such as "a chest radiograph with upper lobe cavitary lesions." These prompts are fed into the VLM's text encoder to generate text embeddings. The embeddings for each class are then averaged and normalized to create a single representative vector, or "text prototype," for that class.

\subsubsection{Zero-Shot Classification}
During inference, a test image is passed through the VLM's image encoder to produce an image embedding. We then compute the cosine similarity between this image embedding and the two text prototypes. The class corresponding to the prototype with the higher similarity score is selected as the prediction. This standard approach is referred to as "argmax" classification.

\subsection{Evaluation Metrics}
To ensure a comprehensive comparison, we evaluate all models using three standard metrics:
\begin{itemize}
    \item \textbf{Accuracy (ACC):} The proportion of correctly classified images.
    \item \textbf{F1-Score (F1):} The harmonic mean of precision and recall, providing a robust measure for binary classification.
    \item \textbf{Area Under the Receiver Operating Characteristic Curve (ROC AUC):} A threshold-independent metric that evaluates the model's ability to distinguish between the two classes.
\end{itemize}
\subsection{Remedy: Threshold Calibration}
A key part of our methodology is to address the sub-optimal performance of the standard `argmax` approach in zero-shot classification. We propose a calibration step that optimizes the decision threshold on a held-out validation set to maximize the F1-score, a metric that balances precision and recall.

The process is as follows. First, for each image $\mathbf{x}_i$ in the validation set $\mathcal{V}$, we compute the VLM's softmax probability $p_i$ for the positive class (pneumonia or TB). Then, for any given probability threshold $\tau \in [0, 1]$, a prediction can be made using the decision rule:
\[
\hat{y}_i(\tau) = \mathbb{1}\{p_i \ge \tau\}.
\]
Our goal is to find the optimal threshold, $\tau^*$, that yields the highest F1-score when comparing the predictions $\{\hat{y}_i(\tau)\}$ against the true validation labels $\{y_i\}$. This is formulated as an optimization problem:
\[
\tau^* = \arg\max_{\tau \in \mathcal{T}} \text{F1}\big(\hat{y}(\tau), y_{\text{val}}\big),
\]
where $\mathcal{T}$ is a dense grid of candidate thresholds (e.g., $[0.02, 0.98]$). This calibrated threshold $\tau^*$ is then used to make final predictions on the held-out test set. It is important to note that this procedure only finds a better operating point for classification; it does not alter the model's underlying discriminative ability, and thus threshold-independent metrics like ROC AUC remain unchanged.

\section{Experiments}

\subsection{Experiment Setup}
\textbf{Datasets:} We use \emph{PneumoniaMNIST} from MedMNIST v2 (loaded in code as \texttt{medmnist.PneumoniaMNIST}) and the \emph{Shenzhen Chest X-Ray Set} (referenced in our codebase as \texttt{ShenzhenTB}). We keep the official split for PneumoniaMNIST (4{,}708/524/624) and a fixed-seed stratified 60\%/10\%/30\% split for Shenzhen.

\textbf{Models:} The supervised baseline is a lightweight CNN (Table~\ref{tab:cnn_arch}; three \mbox{$3{\times}3$} conv–ReLU–maxpool blocks followed by a two-layer MLP head) trained with cross-entropy loss using AdamW (learning rate $10^{-3}$, weight decay $10^{-4}$). Inputs are single-channel $64{\times}64$ grayscale with light augmentation (random horizontal flip and small affine transforms).  The zero-shot model is \emph{BiomedCLIP} (image encoder: ViT-B/16; text encoder as released by the authors). Images are preprocessed by the BiomedCLIP pipeline (RGB, $224{\times}224$). For each class we form five text prompts, encode them, average to a normalized class prototype, and classify by cosine similarity (argmax).

\subsection{Results}
The performance of the models on the test sets of both datasets is summarized in Table \ref{tab:results}. For the VLMs, the calibration procedure was performed on the validation set of each respective dataset. This yielded an optimal F1-score threshold of \(t^*=0.076\) for BiomedCLIP on PneumoniaMNIST and \(t^*=0.020\) on the Shenzhen TB dataset.

\begin{table}[htbp]
\caption{Metrics Comparison on Test Sets}
\begin{center}
\begin{tabular}{@{}lccc@{}}
\toprule
\textbf{Model} & \textbf{ACC} & \textbf{F1} & \textbf{ROC AUC} \\
\midrule
\multicolumn{4}{l}{\textit{PneumoniaMNIST (Pneumonia Detection)}} \\
CNN (trained)                      & 0.8317 & 0.8803 & 0.9314 \\
BiomedCLIP (argmax)     & 0.7660 & 0.7747 & 0.9228 \\
BiomedCLIP (calibrated) & \textbf{0.8542} & \textbf{0.8841} & 0.9228 \\
\midrule
\multicolumn{4}{l}{\textit{Shenzhen (Tuberculosis Detection)}} \\
CNN (trained)                      & 0.7638 & \textbf{0.7834} & \textbf{0.8755} \\
BiomedCLIP (argmax)     & 0.6533 & 0.4812 & 0.8569 \\
BiomedCLIP (calibrated) & \textbf{0.7789} & 0.7684 & 0.8569 \\
\bottomrule
\end{tabular}
\label{tab:results}
\end{center}
\end{table}
\subsection{Computational Efficiency}
To contextualize accuracy, we compare model size and single-image inference time on the same device. The CNN has \textbf{285{,}634} trainable parameters (Table~\ref{tab:cnn_arch}), whereas BiomedCLIP couples an image encoder and a text encoder and is typically two to three orders of magnitude larger (for example, \texttt{ViT-B/16} or \texttt{ViT-L/14} backbones yield hundreds of millions of parameters). This highlights the performance–efficiency trade-off: the calibrated zero-shot VLM narrows or exceeds the CNN on F1 while requiring substantially greater compute and memory.

\subsection{Analysis and Discussion}
Our results provide several key insights across both clinical tasks. First, the supervised CNNs establish very strong baselines, achieving F1-scores of 0.8803 for pneumonia and 0.7834 for TB. This demonstrates that for well-defined tasks with sufficient labeled data, a lightweight, specialized CNN can be highly effective.

Second, the out-of-the-box zero-shot performance of BiomedCLIP is respectable but does not surpass the supervised baselines. Using the standard argmax method, the F1-scores are significantly lower than the CNNs on both datasets (0.7747 vs. 0.8803 for pneumonia; 0.4812 vs. 0.7834 for TB).

Most importantly, our proposed calibration remedy has a significant and consistently positive impact. For \textbf{BiomedCLIP on PneumoniaMNIST}, calibrating the decision threshold provides a dramatic performance boost, increasing the F1-score from 0.7747 to 0.8841. This data-driven adjustment allows the zero-shot model to not only match but exceed the performance of the fully supervised CNN.

On the \textbf{Shenzhen TB dataset}, the effect of calibration is even more pronounced. It elevates the F1-score from a poor 0.4812 to a very competitive 0.7684. While this calibrated score is slightly below the supervised CNN's F1-score of 0.7834, it represents a massive improvement that makes the zero-shot model viable for the task. This result highlights that while calibration may not always guarantee superiority, it is a critical step to unlock the VLM's inherent capabilities.

Finally, it is worth noting that the ROC AUC scores for the VLM remain constant regardless of calibration. This is expected, as ROC AUC is a threshold-independent metric. The high AUC scores (0.9228 and 0.8569) indicate that the VLM has strong underlying discriminative power on both tasks, even in a zero-shot setting. Our remedy simply finds a better operating point for making discrete predictions, which is crucial for practical applications.

\subsection{Qualitative Analysis}
To better understand the models' behaviors, we conduct a qualitative analysis of their predictions.

\begin{figure*}[t]
\centerline{\includegraphics[width=0.9\textwidth]{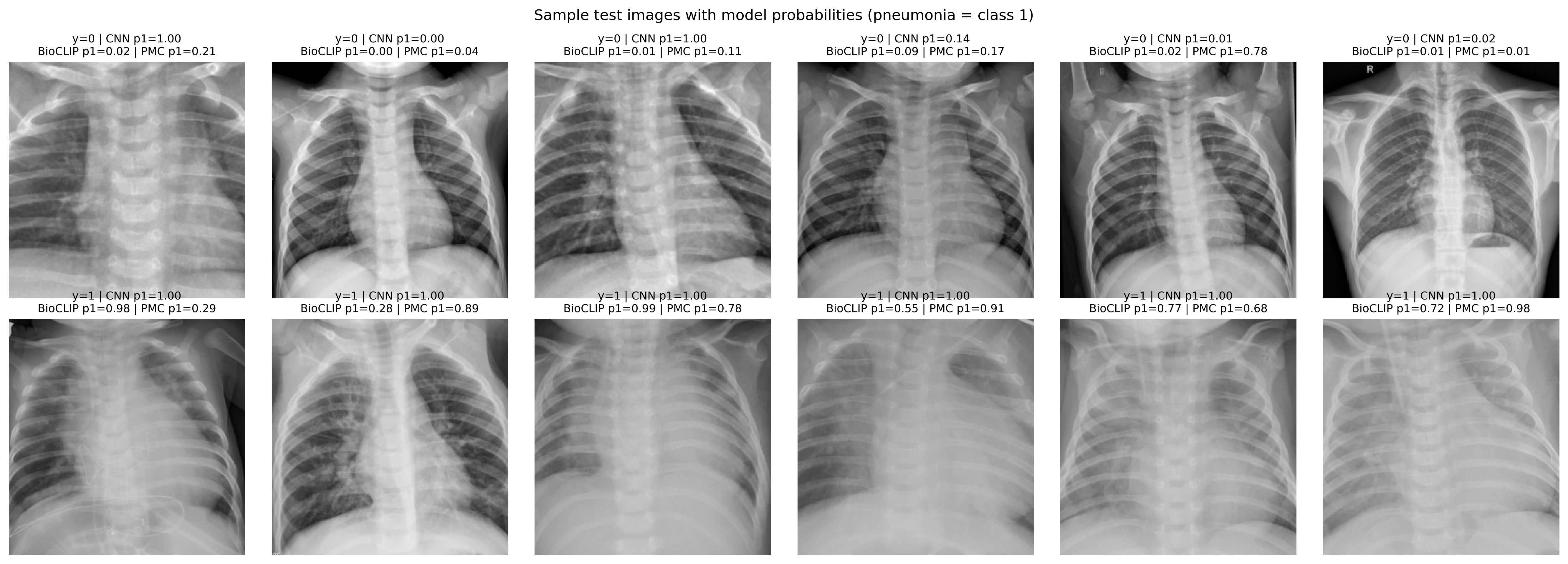}}
\caption{Sample test images from PneumoniaMNIST with predicted probabilities for the pneumonia class (p1) from the trained CNN and BiomedCLIP (BioCLIP). The top row shows normal cases (y=0) and the bottom row shows pneumonia cases (y=1). \textbf{All probability color maps use a common 0--1 legend for visual comparability.}}

\label{fig:qual_panel}
\end{figure*}

Figure \ref{fig:qual_panel} displays sample images from the PneumoniaMNIST test set. The trained CNN is highly confident in its predictions, with probabilities typically being either 1.00 or 0.00. In contrast, the zero-shot VLM provides more graded probabilities, which helps explain why threshold calibration is so impactful.

\begin{figure*}[t]
\centerline{\includegraphics[width=0.8\textwidth]{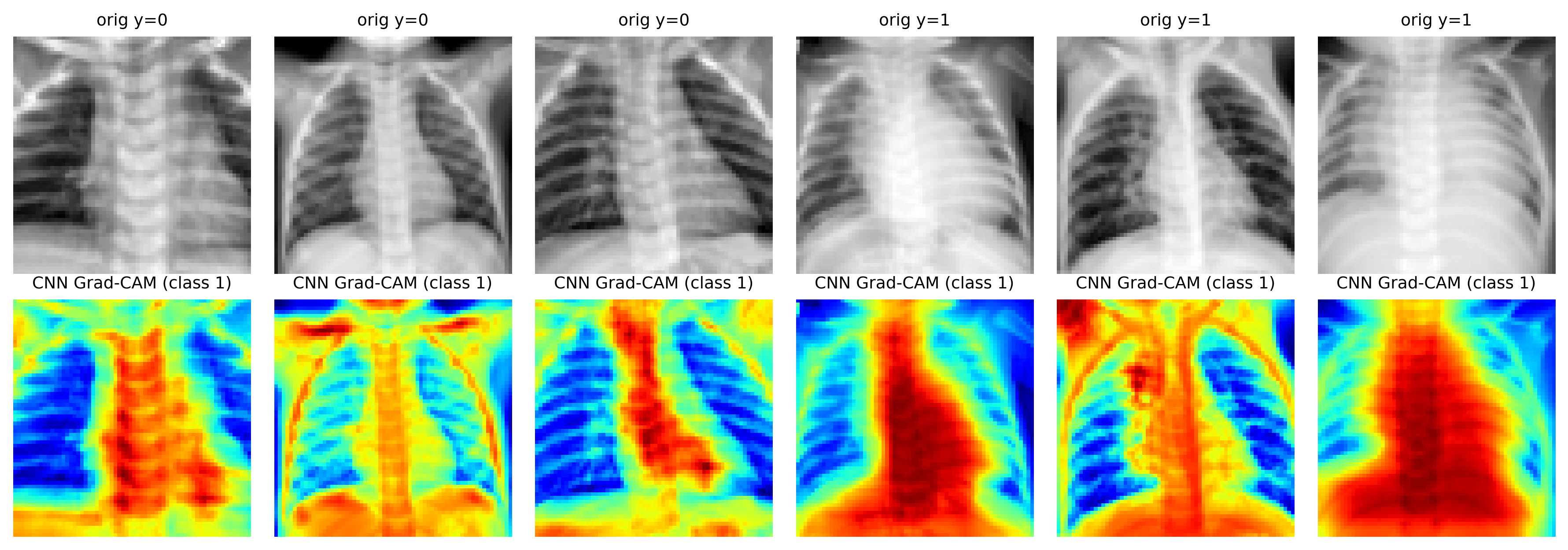}}
\caption{Grad-CAM visualizations for the CNN trained on PneumoniaMNIST. The heatmaps (bottom row) highlight the image regions most influential for predicting pneumonia. Red indicates high importance. The model correctly focuses on lung fields in pneumonia cases (right). \textbf{This alignment between Grad-CAM heat maps and lung fields supports the clinical plausibility of CNN predictions.}}

\label{fig:gradcam}
\end{figure*}

To investigate the interpretability of our strong CNN baseline, we use Gradient-weighted Class Activation Mapping (Grad-CAM) on the pneumonia detection task. As shown in Figure \ref{fig:gradcam}, the results are clinically relevant. For true pneumonia cases (right), the model's attention (red areas) is concentrated on the lung fields, where radiographic evidence of pneumonia would be expected. For normal cases (left), the activation is more diffuse. This provides evidence that the CNN has learned a valid, feature-based strategy for classification.

\subsection{Limitations}
This study has several limitations. First, we evaluate only two binary chest X-ray tasks, which may not represent broader clinical variability. Second, dataset size and class imbalance can bias threshold selection and F1. Third, zero-shot prompts may not capture the full range of radiological phrasing, which can depress VLM probabilities; richer prompt sets or prompt ensembling could help. Finally, images are resized and preprocessed for modeling, so physical scale is not preserved in qualitative panels.

\section{Conclusion and Future Work}
In this work, we conducted a comparative study of a supervised, lightweight CNN and a zero-shot medical VLM for chest X-ray classification across two tasks: pneumonia and tuberculosis detection. We demonstrated that supervised CNNs provide powerful and efficient baselines. Our key finding is that the default zero-shot application of VLMs can be misleadingly suboptimal. By introducing a  decision threshold calibration step, we were able to significantly improve the performance of BiomedCLIP on both datasets. This remedy enabled the VLM to surpass the supervised baseline for pneumonia detection and to become highly competitive for tuberculosis detection. This underscores the critical importance of post-processing and calibration when deploying large pre-trained models in specialized domains like medical imaging.
In future work, we will explore calibration methods beyond a simple threshold search (e.g., temperature scaling) and study a few-shot setting where VLMs are fine-tuned with \textbf{1\%, 5\%, and 10\%} of labeled training images to quantify data–performance curves. We will investigate federated learning on decentralized data using selective layer fine-tuning \cite{zhang2025selective} and privacy-preserving attention \cite{li2025selective}. To assess generalizability, we will apply this comparison framework to additional medical imaging tasks and datasets. We will also extend our calibrated zero-shot setup to infectious-disease and barrier-function problems informed by recent studies from Li and collaborators \cite{li2025pnas,hu2025interplay,wang2025gos,li2024nsp1,chanez2024mlck,li2023pparg,li2021archtox,li2021faseb} and to physics-informed and graph-enhanced Lattice Boltzmann modeling for multiphase and turbulent flows to generate mechanistic augmentations and test calibration robustness \cite{li2025physics,li2025lbm,li2025spectral}. Finally, we will prioritize \textbf{explainability and trust}, including Grad-CAM auditing protocols and reader-study designs, to support clinical adoption.


\begin{thebibliography}{00}
\bibitem{b1} J. Yang, R. Shi, and D. Ni, "MedMNIST v2: A large-scale lightweight benchmark for 2D and 3D biomedical image classification," \textit{Scientific Data}, vol. 10, no. 1, p. 41, 2023.
\bibitem{b2} S. Zhang, Z. Wu, V. K. Singh, and A. G. Schwing, "Large-scale domain-specific pretraining for biomedical vision-language processing," in \textit{Proceedings of the IEEE/CVF Winter Conference on Applications of Computer Vision}, 2024, pp. 8399-8409.
\bibitem{b3} R. Yip, "PMC-CLIP: Contrastive Language-Image Pre-training using Biomedical Documents," in \textit{Machine Learning for Health (ML4H)}, 2023.
\bibitem{b4} R. Tong, T. Xu, X. Ju, and L. Wang, "Progress in Medical AI: Reviewing Large Language Models and Multimodal Systems for Diagonosis," \textit{AI Med}, vol. 1, no. 1, p. 5, 2025.
\bibitem{b5} J.~Liu, R.~Tong, A.~Shen, S.~Li, C.~Yang, and L.~Xu, ``MemeBLIP2: A novel lightweight multimodal system to detect harmful memes,'' \emph{arXiv:2504.21226} [cs.CV], 2025. doi:10.48550/arXiv.2504.21226.
\bibitem{b6} R.~Tong, L.~Wang, T.~Wang, and W.~Yan, ``Predicting Parkinson's Disease Progression Using Statistical and Neural Mixed Effects Models: A Comparative Study on Longitudinal Biomarkers,'' \emph{arXiv:2507.20058} [stat.ML], 2025. doi:10.48550/arXiv.2507.20058.
\bibitem{b7} G. Litjens et al., "A survey on deep learning in medical image analysis," \textit{Medical Image Analysis}, vol. 42, pp. 60-88, 2017.
\bibitem{b8} A. Radford et al., "Learning transferable visual models from natural language supervision," in \textit{International Conference on Machine Learning}, PMLR, 2021, pp. 8748-8763.
\bibitem{shenzhen_qims}
S. Jaeger, S. Candemir, S. Antani, Y.-X. J. Wáng, P.-X. Lu, and G. Thoma,
``Two public chest X-ray datasets for computer-aided screening of pulmonary tuberculosis,''
\textit{Quantitative Imaging in Medicine and Surgery}, vol. 4, no. 6, pp. 475--477, 2014.
doi:10.3978/j.issn.2223-4292.2014.11.20.
\bibitem{zhu2018nonlinear}
X. Zhu, X. Shen, X. Jiang, K. Wei, T. He, Y. Ma, J. Liu, and X. Hu, "Nonlinear expression and visualization of nonmetric relationships in genetic diseases and microbiome data," \textit{BMC bioinformatics}, vol. 19, pp. 37--48, 2018.
\bibitem{li2025ammkd}
Y. Li, C. Yang, J. Dong, Z. Yao, H. Xu, Z. Dong, H. Zeng, Z. An, and Y. Tian, "AMMKD: Adaptive Multimodal Multi-teacher Distillation for Lightweight Vision-Language Models," \textit{arXiv preprint arXiv:2509.00039}, 2025.

\bibitem{li2025achieving}
Y. Li, Y. Li, K. Zhang, F. Zhang, C. Yang, Z. Guo, W. Ding, and T. Huang, "Achieving Fair Medical Image Segmentation in Foundation Models with Adversarial Visual Prompt Tuning," \textit{Information Sciences}, pp. 122501, 2025.
\bibitem{liu2025markuplanguagemodelingweb}
S. Liu, B. Bi, J. Bakus, P. K. Velalam, V. Yella, and V. Hegde, "Markup Language Modeling for Web Document Understanding," \textit{arXiv preprint arXiv:2509.20940}, 2025.
\bibitem{zhang2025selective}
L. Zhang and Y. Li, "Selective Layer Fine-Tuning for Federated Healthcare NLP: A Cost-Efficient Approach," in \textit{2025 International Conference on Artificial Intelligence, Computer, Data Sciences and Applications (ACDSA 2025)}, 2025.

\bibitem{li2025selective}
Y. Li and L. Zhang, "Selective Attention Federated Learning: Improving Privacy and Efficiency for Clinical Text Classification," in \textit{2025 International Conference on Artificial Intelligence, Computer, Data Sciences and Applications (ACDSA 2025)}, 2025.

\bibitem{wang2025systematic}
Y. Wang, J. Zhong, and R. Kumar, "A systematic review of machine learning applications in infectious disease prediction, diagnosis, and outbreak forecasting," \textit{Preprints}, 2025.

\bibitem{wang2025applications}
Y. Wang, Z. Wang, J. Zhong, D. Zhu, and W. Li, "Applications of Small Language Models in Medical Imaging Classification with a Focus on Prompt Strategies," \textit{arXiv preprint arXiv:2508.13378}, 2025.
\bibitem{li2025pnas}
E.~Li, R.~Zang, T.~Kawagishi, W.~Zhang, K.~Iyer, G.~Hou, Q.~Zeng, \emph{et al.},
“Fatty acid 2-hydroxylase facilitates rotavirus uncoating and endosomal escape,”
\emph{Proceedings of the National Academy of Sciences of the USA}, vol.~122, no.~36, article e2511911122, 2025.

\bibitem{hu2025interplay}
W.~Hu, Y.~Qiao, E.~Li, M.~Li, and L.~Che,
“Interplay between nutrition, microbiota, and immunity in rotavirus infection: insights from human and animal models,”
\emph{Frontiers in Veterinary Science}, vol.~12, article 1680448, 2025.

\bibitem{wang2025gos}
Y.~Wang, Z.~Li, G.~Chen, Y.~Xing, J.~Wang, Y.~Zhao, M.~Kang, K.~Huang, E.~Li, B.~Tan, \emph{et al.},
“Dietary galacto-oligosaccharides enhance growth performance and modulate gut microbiota in weaned piglets: a sustainable alternative to antibiotics,”
\emph{Animals}, vol.~15, no.~11, article 1508, 2025.

\bibitem{li2024nsp1}
E.~Li, N.~Feng, Q.~Zeng, L.~Sanchez-Tacuba, T.~Kawagishi, G.~Branham, \emph{et al.},
“Rhesus rotavirus NSP1 mediates extra-intestinal infection and is a contributing factor for biliary obstruction,”
\emph{PLOS Pathogens}, vol.~20, no.~9, article e1016209, 2024.

\bibitem{chanez2024mlck}
O.~Chanez-Perandres, S.~Abtahi, J.~Zha, E.~Li, G.~Li, S.~Prasitsilp, L.~Zuo, M.~J.~Grey, \emph{et al.},
“Mechanisms underlying distinct subcellular localization and regulation of epithelial long myosin light-chain kinase splice variants,”
\emph{Journal of Biological Chemistry}, 2024. 

\bibitem{li2023pparg}
E.~Li, C.~Li, N.~Horn, and K.~M.~Ajuwon,
“PPAR$\gamma$ activation inhibits endocytosis of claudin-4 and protects against deoxynivalenol-induced intestinal barrier dysfunction in IPEC-J2 cells and weaned piglets,”
\emph{Toxicology Letters}, vol.~375, pp.~8--20, 2023. 
\bibitem{xiao2025depthssc}
J.~Yao, J.~Zhang, X.~Pan, T.~Wu, and C.~Xiao, ``DepthSSC: Monocular 3D semantic scene completion via depth–spatial alignment and voxel adaptation,'' in \textit{Proc. IEEE/CVF Winter Conference on Applications of Computer Vision (WACV)}, 2025.

\bibitem{yao2024swiftsampler}
J.~Yao, C.~Li, and C.~Xiao, ``Swift sampler: Efficient learning of sampler by 10 parameters,'' \textit{Advances in Neural Information Processing Systems}, vol.~37, 2024.


\bibitem{li2021archtox}
E.~Li, N.~Horn, and K.~M.~Ajuwon,
``Mechanisms of deoxynivalenol-induced endocytosis and degradation of tight junction proteins in jejunal IPEC-J2 cells involve selective activation of the MAPK pathways,''
\emph{Archives of Toxicology}, vol.~95, no.~6, pp.~2065--2079, 2021.

\bibitem{li2021faseb}
E.~Li and K.~M.~Ajuwon,
``Mechanism of endocytic regulation of intestinal tight junction remodeling during nutrient starvation in jejunal IPEC-J2 cells,''
\emph{The FASEB Journal}, vol.~35, no.~2, e21356, 2021.

\bibitem{su2025applications}
Y.~H.~Su, Z.~Y.~Lu, J.~H.~Liu, K.~Pang, H.~R.~Dai, S.~Liu, Y.~X.~Jia, L.~J.~Ge, and J.~M.~Yang,
``Applications of large models in medicine,''
\emph{arXiv:2502.17132}, 2025.

\bibitem{zhang2015thalamocortical}
C.~H.~Zhang, Z.~Sha, J.~Mundahl, S.~Liu, Y.~F.~Lu, T.~R.~Henry, and B.~He,
``Thalamocortical relationship in epileptic patients with generalized spike and wave discharges: A multimodal neuroimaging study,''
\textit{NeuroImage: Clinical}, vol.~9, pp.~117--127, 2015.
\bibitem{li2025physics}
Y.~Li, ``Physics-Informed Neural Networks for Enhanced Interface Preservation in Lattice Boltzmann Multiphase Simulations,'' in \textit{Proc. 2025 International Conference on Artificial Intelligence, Computer, Data Sciences and Applications (ACDSA 2025)}, 2025.

\bibitem{li2025lbm}
Y.~Li, ``LBM\textendash GNN: Graph Neural Network Enhanced Lattice Boltzmann Method,'' in \textit{Proc. 2025 International Conference on Artificial Intelligence, Computer, Data Sciences and Applications (ACDSA 2025)}, 2025.

\bibitem{li2025spectral}
Y.~Li, S.~Liu, and Z.~Xu, ``Spectral Decomposition PINN\textendash LBM for High Reynolds Number Turbulence Simulation,'' in \textit{Proc. 9th Computational Methods in Systems and Software (CoMeSySo 2025)}, Springer, 2025.

\end{thebibliography}
\end{document}